\documentclass[sigconf,nonacm]{acmart}
\AtBeginDocument{%
  \providecommand\BibTeX{{%
    \normalfont B\kern-0.5em{\scshape i\kern-0.25em b}\kern-0.8em\TeX}}}

\renewcommand\footnotetextcopyrightpermission[1]{} % removes footnote with conference information in first column
\setcopyright{none}

\usepackage{graphicx}
\usepackage{amsmath}
\usepackage{booktabs}
\usepackage{amsmath, bm, epstopdf, url, pifont, overpic, cases}
\usepackage{url}
\usepackage{booktabs}
\usepackage{bbding}
\usepackage{pifont}
\usepackage{wasysym}
\usepackage{utfsym}
\usepackage{fontawesome}
\usepackage{multirow}
\usepackage{algorithm}
\usepackage{algorithmic}
\usepackage{tabularx}
\usepackage{multirow}
\usepackage {pifont}
\usepackage{colortbl} % 引入colortbl包
\usepackage{xcolor} % 引入xcolor包

\usepackage{bbding}
\usepackage{pifont}
\usepackage{wasysym}

\usepackage{arydshln}
\usepackage{nicefrac}

\usepackage{colortbl}
\usepackage{xcolor}
\usepackage{color}
\usepackage{array}  
\usepackage{enumitem}
\usepackage{xfrac} 
\usepackage{makecell}
\usepackage{url}
\usepackage[misc]{ifsym}

\usepackage{colortbl}
\usepackage{xcolor}
\usepackage{array}  
\definecolor{color3}{rgb}{0.95,0.95,0.95}
\usepackage{pifont}

\usepackage{fontawesome}
\usepackage[capitalize]{cleveref}
\crefname{section}{Sec.}{Secs.}
\Crefname{section}{Section}{Sections}
\Crefname{table}{Table}{Tables}
\crefname{table}{Tab.}{Tabs.}

\begin{document}

\title{Diffusion-Based Hierarchical Image Steganography}

% \affiliation{%

\author{
Youmin Xu$^{1, 2}$, Xuanyu Zhang$^{1, 3}$, Jiwen Yu$^{1}$, Chong Mou$^{1}$, Xiandong Meng$^{2}$, Jian Zhang$^{1, 3 }$\textsuperscript{\faEnvelopeO}\\
$^1$School of Electronic and Computer Engineering, Peking University\\
$^2$Peng Cheng Laboratory\\
$^3$Peking University Shenzhen Graduate School-Rabbitpre AIGC Joint Research Laboratory
}
%% You do not have to enter your paper ID



\begin{abstract}
This paper introduces Hierarchical Image Steganography (HIS), a novel method that enhances the security and capacity of embedding multiple images into a single container using diffusion models. HIS assigns varying levels of robustness to images based on their importance, ensuring enhanced protection against manipulation. It adeptly exploits the robustness of the Diffusion Model alongside the reversibility of the Flow Model. The integration of Embed-Flow and Enhance-Flow improves embedding efficiency and image recovery quality, respectively, setting HIS apart from conventional multi-image steganography techniques. This innovative structure can autonomously generate a container image, thereby securely and efficiently concealing multiple images and text. Rigorous subjective and objective evaluations underscore HIS's advantage in analytical resistance, robustness, and capacity, illustrating its expansive applicability in content safeguarding and privacy fortification.
\end{abstract}






\begin{teaserfigure}
\centering
\includegraphics[width=0.94\linewidth]{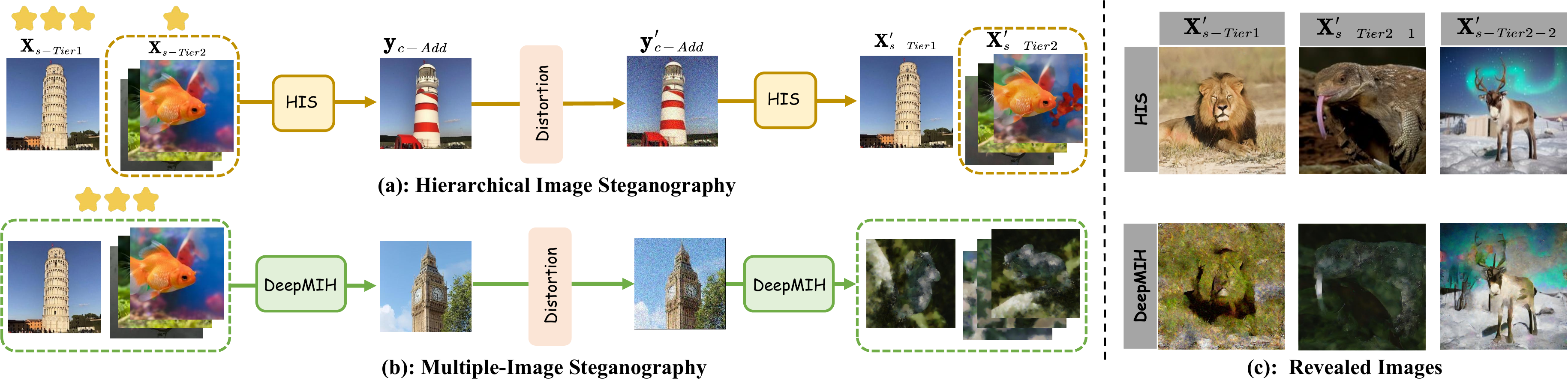}
% \captionsetup{type=figure}
\caption{Workflow and results of (a)proposed HIS and (b)former DeepMIH \cite{guan2022deepmih}: HIS hides more important $\mathbf{x}_{s-Tier1}$ and less important $\mathbf{x}_{s-Tier2}$ into a generated container $\mathbf{y}_{c-Add}$. Compared to $\mathbf{x}'_{s-Tier2}$, $\mathbf{x}^{\prime}_{s-Tier1}$ can be more robustly recovered from distorted $\mathbf{y}'_{c-Add}$ by the receiver. Multiple Image Steganography \cite{guan2022deepmih} usually treats every secret image with the same priority and ignores the robustness, resulting in the worse reconstruction over all images under distortion. (c) presents more results of revealed secret images by HIS and \cite{guan2022deepmih}.}
% \Description{figure description}
\label{fig:teaser} 
\vspace{-1pt}
\end{teaserfigure}

\maketitle

 % Here the input data is binarized and further mapped into the color-quantized host image. 

\section{Introduction}
\label{intro}

%While deep image steganography achieves compelling performance on visual quality and capacity and image benchmarks

% the hiding network and the revealing network are independent of each other. 
% the hiding network and the revealing network in deep image steganography are all jointly optimized, in which the floating-point images flow through both networks for forward and backward propagations.

\begin{figure*}[t]
    \centering
\includegraphics[width=0.88\linewidth]{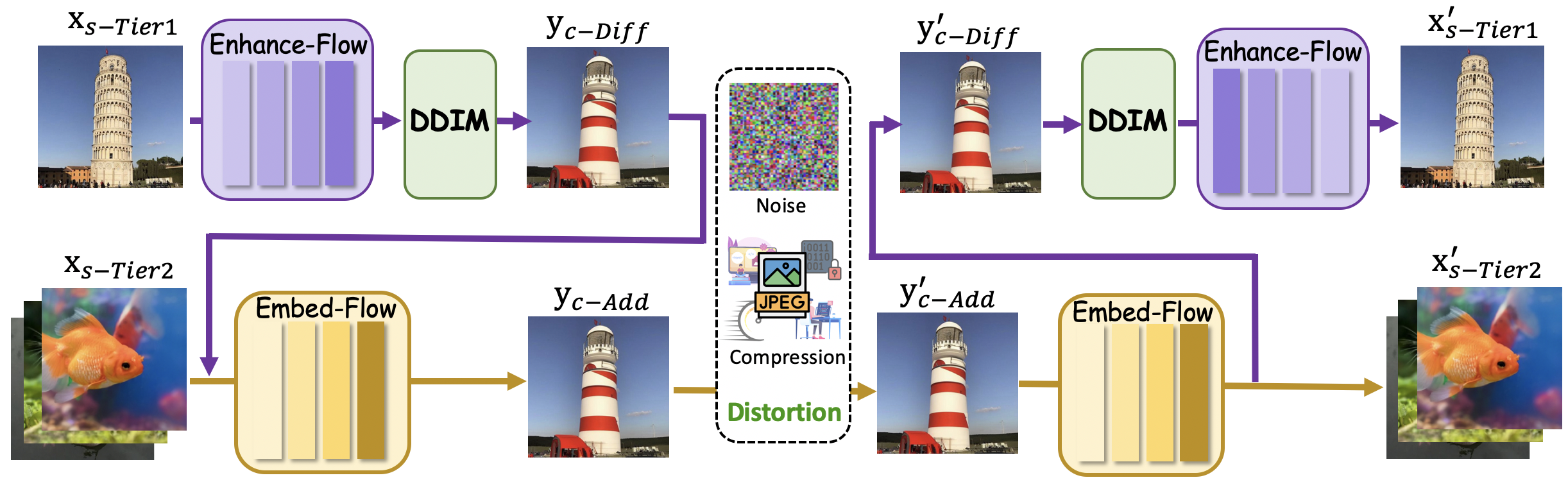}
    \caption{Framework of our diffusion-based Hierarchical Image Steganography (HIS), which takes $\mathbf{x}_{s-Tier1}$ and multiple $\mathbf{x}_{s-Tier2}$ as input to produced $\mathbf{y}_{c-Add}$ as container. Inversely, $\mathbf{x}^{\prime}_{s-Tier1}$ is recovered via Enhance-Flow and $\mathbf{x}^{\prime}_{s-Tier2}$ is recovered via Embed-Flow from container.
    }
    \vspace{-0.1cm}
    \label{fig:frame}
\end{figure*}
Steganography is a widely studied topic \cite{baluja2017hiding, baluja2019hiding}, which aims to hide messages like audio, image, and text into one container image in an undetected manner. In its reverse process, it is only possible for the receivers with a correct revealing network to reconstruct secret information from the container, which is visually identical to the host. For image steganography, traditional methods often use adaptive encoding according to distortion cost designed by human or neural networks \cite{baluja2017hiding, baluja2019hiding, jing2021hinet}, which require rules and knowledge.

%With the development of deep neural networks, researchers begin to use auto-encoder networks \cite{ddh,ddh_2,udh,beforeHidden} or invertible neural networks(INN) \cite{ISN,HiNet} to hide data, which is called deep steganography. They show great success in visual quality, but can't guarantee accurate extraction. It is because some works in deep steganography aim to hide secret images within images\cite{ddh,ddh_2,udh,ISN,HiNet}:

% \begin{figure*}[t]
%   \centering
%   \vspace{-0.1cm}
%   \includegraphics[width=1\linewidth]{figure/method-cross-v2.pdf}
%   \vspace{-0.5cm}
%   \caption{Our coverless image steganography framework CRoSS. The diffusion model we choose is a conditional diffusion model, which supports conditional inputs to control the generation results. We choose the deterministic DDIM as the sampling strategy and use the two different conditions ($\mathbf{k}_{pri}$ and $\mathbf{k}_{pub}$) given to the model as the private key and the public key.}
%   % \vspace{-0.4cm}
% \label{fig:cross_method} 
% \end{figure*}

% \begin{figure*}[t]
%   \centering
%   \vspace{-0.1cm}
%   \includegraphics[width=1\linewidth]{figure/prompt.pdf}
%   \vspace{-0.5cm}
%   \caption{Visual results of the proposed CRoSS controlled by different prompts. The container images are realistic and the revealed images have well semantic consistency with the secret images.}
%   % \vspace{-0.4cm}
% \label{fig:prompt} 
% \end{figure*}4

In the realm of image steganography, there are four primary objectives to consider: \textbf{payload capacity, undetectability, integrity}, and \textbf{robustness}. Steganalysis methodologies typically differentiate between the host and standard images using attributes like color, frequency, and other distinctive features. It's thus imperative that the concealed image is embedded within the unseen spectrum of the host image. Since the earlier steganography methods stress capacity and invisibility rather than robustness and ignore the noise and compression interference in practice, they are usually sensitive to distortion during the transmission of the container.  

Traditional approaches to \textbf{\textit{Multiple Image Steganography}}\cite{xu2022robust, lu2021large}, or to say \textbf{\textit{Multiple Image Hiding}}\cite{guan2022deepmih}, involve embedding multiple images into a single container image without differentiating the importance of each image. Consequently, if the container image is  degraded, all embedded information risks being lost. Recognizing the practical necessity in real-world applications, some images demand higher robustness to maintain integrity post-degradation, while others may tolerate some compromise. To address this, we introduce the concept of \textbf{\textit{Hierarchical Steganography}}, which tailors embedding robustness according to the importance of the secret content. This tailored scheme represents an adaptive innovation in the field of steganography. 

Guided by this novel design philosophy, we propose the \textbf{Diffusion-Based Hierarchical Image Steganography (HIS)}, advancing beyond existing methods by robustness, security and capacity. It implements a tiered embedding strategy that effectively manages the robustness of various images embedded within a single, model-generated container. 
Recently, diffusion models have been widely used in various image applications, including image generation~\cite{dhariwal2021diffusion,ramesh2022hierarchical,saharia2022photorealistic,rombach2021highresolution}, restoration~\cite{saharia2022image,kawar2022denoising,wang2023zeroshot}, translation~\cite{Choi_2021_ICCV,Kim_2022_CVPR,meng2022sdedit,zhao2022egsde}, and more. HIS incoporates the diffusion model as the core of generation from secret images to container image. In Fig.\ref{fig:teaser}, HIS differentiates itself by embedding images according to their importance: Tier-1 images are embedded with the highest robustness to ensure their integrity against distortions, while Tier-2 images are embedded with lower robustness. This novel strategy not only maximizes the embedding capacity of the container image but also introduces a dynamic content prioritization mechanism that can be tailored according to specific security requirements.

% The Stable Diffusion~\cite{rombach2021highresolution} community is currently one of the most popular and thriving ones, with a large number of open-source tools available for free, including model checkpoints finetuned on various specialized datasets. Additionally, various LoRAs~\cite{hu2022lora} and ControlNets~\cite{zhang2023adding} are available in these communities for efficient control over the results generated by Stable Diffusion. LoRAs achieve control by efficiently modifying network parameters in a low-rank way, while ControlNets introduce an additional network to modify the intermediate features of Stable Diffusion for control. These mentioned recent developments have enhanced our Diff-StegFlow.

The main contributions are summarized as follows:

\begin{itemize}

\item By recognizing the need to assign different levels of robustness based on the importance of content, the scheme of \textit{Hierarchical Steganography} innovatively addresses the limitations of traditional multi-image steganography.

\item We propose diffusion-based Hierarchical Image Steganography (HIS), which hides large-capacity images within undetectable containers as substitute and enhances the flexibility of steganography in practical applications.

\item HIS incorporates Embed-Flow and Enhance-Flow, which significantly improve the steganographic capacity and recovery quality of diffusion. Owing to the reversible characteristics of the flow model, the result of diffusion-based inversion is guided toward more deterministic results, which is also enlightening for applications of diffusion models.

\item  HIS method outperforms existing \textit{Multi Image Steganography} techniques across several critical metrics, including security, robustness, and capacity. This advancement sets a new benchmark for the efficacy and applicability of steganographic practices in securing digital content.

% \item A robust and comprehensive framework is proposed for coverless image steganography. Compared to previous similar methods.

% \item We introduced the concept of importance-based steganography for the first time. The resulting container images can hide large-capacity image and text information within typical minor degradation ranges and still preserve the vital secret images under severe degradation, such as high compression, screen capture, and printing.

% \item This paper only needs to train a flow model, using it as a supplement to existing diffusion models. Owing to the reversible characteristics and distribution transformation of the flow model, the restoration of coverless steganography is guided toward more deterministic results, which is also enlightening for other applications of diffusion models.

% \item Experimental evidence demonstrates that the coverless steganography proposed in this paper surpasses existing image steganography methods in terms of maximum steganographic capacity, robustness, and security against detection analysis.

\end{itemize}

\begin{figure*}[t]
 % \vspace{-5pt}
  \centering
 
 % \fbox{\rule{0pt}{0.5in} \rule{0.9\linewidth}{0pt}}
  \includegraphics[width=0.86\linewidth]{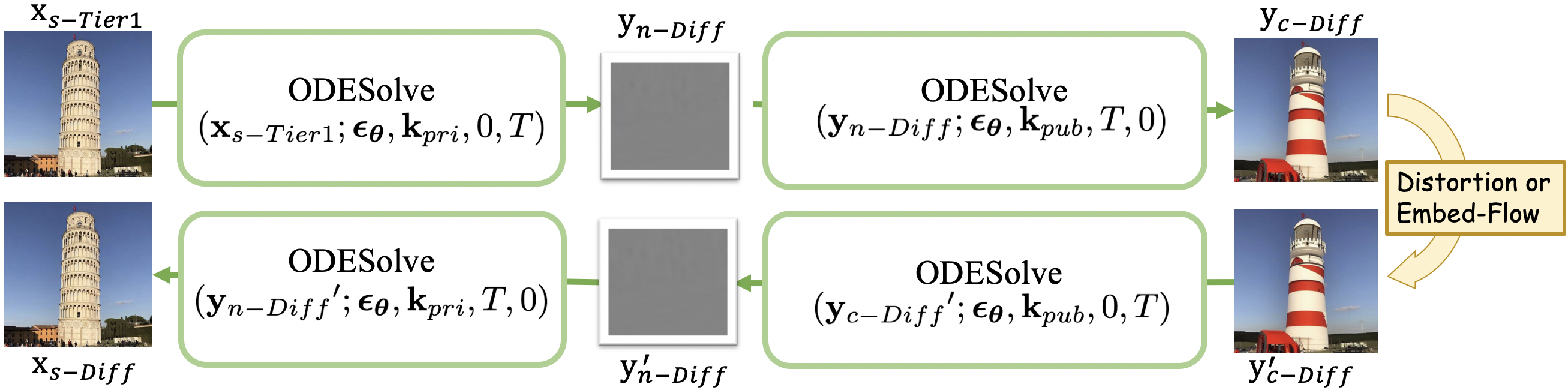}
  \vspace{-5pt}
  \caption{The workflow of Diffusion-based Inversion: $\mathbf{x}_{s-Tier1}$ is firstly transformed to intermedia noise $\mathbf{y}_{n-Diff}$ with private key $\mathbf{k}_{pri}$, and then transformed to container $\mathbf{y}_{c-Diff}$ with public key $\mathbf{k}_{pub}$ using the DDIM. Reversely, public key $\mathbf{k}_{pub}$ is initially ultilized to transfer $\mathbf{y}^{\prime}_{c-Diff}$ to $\mathbf{y}^{\prime}_{n-Diff}$. Finally, $\mathbf{x}^{}_{Diff}$ is reconstructed under the condition of $\mathbf{k}_{pri}$.}
   \label{fig:ddim-inversion}
\end{figure*}

\section{Related Works}

\textbf{Cover-based Image Steganography. } Unlike cryptography, steganography aims to hide secret data in a host to produce an information container. For image steganography, a cover image is required to hide the secret image in it~\cite{baluja2017hiding}. 
Traditionally, spatial-based~\cite{nguyen2006multi,pan2011image,provos2003hide} methods utilize the Least Significant Bits (LSB), pixel value differencing (PVD)~\cite{pan2011image}, histogram shifting~\cite{tsai2009reversible}, multiple bit-planes~\cite{nguyen2006multi} and palettes~\cite{niimi2002high} to hide images, which may arise statistical suspicion and are vulnerable to steganalysis methods. Adaptive methods~\cite{pevny2010using,li2014new} decompose the steganography into embedding distortion minimization and data coding, which is indistinguishable by appearance but limited in capacity. Various transform-based schemes~\cite{chanu2012image,kadhim2019comprehensive} including JSteg~\cite{provos2003hide} and DCT steganography~\cite{hetzl2005graph} also fail to offer high payload capacity. Recently, various deep learning-based schemes have been proposed to solve image steganography. Baluja~\cite{baluja2017hiding} proposed the first deep-learning method to hide a full-size image into another image. Generative adversarial networks (GANs)~\cite{shi2017ssgan} are introduced to synthesize container images. Probability map methods focus on generating various cost functions satisfying minimal-distortion embedding~\cite{pevny2010using,tang2017automatic}. \cite{yang2019embedding} proposes a generator based on U-Net. \cite{tang2019cnn} presents an adversarial scheme under distortion minimization. Three-player game methods like SteganoGAN \cite{zhang2019steganogan} and HiDDeN \cite{zhu2018hidden} learn information embedding and recovery by auto-encoder architecture to adversarially resist steganalysis. 
Recent attempts~\cite{xiao2020invertible} to introduce invertible neural networks (INN) into low-level inverse problems like denoising, rescaling, and colorization show impressive potential over auto-encoder, GAN \cite{abdal2019image2stylegan}, and other learning-based architectures. Recently, \cite{lu2021large,guan2022deepmih} 
% HiNet~\cite{hinet} and RIIS~\cite{riis} 
proposed designing the steganography model as an invertible neural network (INN)~\cite{nice,realnvp} to perform image hiding and recovering with a single INN model.

% This work was then extended in~\cite{tpami} by permuting the pixels of the secret image to enhance data security. 

% The previous schemes reveal the potential of image steganography in digital communication, copyright protection, information certification, e-commerce, and many other practical fields \cite{cheddad2010digital}.

\noindent \textbf{Coverless Steganography. } Coverless steganography is an emerging technique in the field of information hiding, which aims to embed secret information within a medium without modifying the cover object \cite{qin2019coverless}. Unlike traditional steganography methods that require a cover medium (e.g., an image or audio file) to be altered for hiding information, coverless steganography seeks to achieve secure communication without introducing any changes to the cover object \cite{li2022coverless}. This makes it more challenging for adversaries to detect the presence of hidden data, as there are no observable changes in the medium's properties \cite{mohamed2021coverless}. To the best of our knowledge, existing coverless steganography methods \cite{liu2020coverless} still focus on hiding bits into container images, and few explorations involve hiding images without resorting to cover images.

\subsection{Diffusion Models}
Diffusion models~\cite{ho2020denoising} are a type of generative model that is trained to learn the target image distribution from a noise distribution. Recently, due to their powerful generative capabilities, diffusion models have been widely used in various image applications, including image generation~\cite{dhariwal2021diffusion}, restoration~\cite{saharia2022image}, translation~\cite{Choi_2021_ICCV}, and more. Large-scale diffusion model communities have also emerged on the Internet, with the aim of promoting the development of AIGC(AI-generated content)-related fields by applying the latest advanced techniques.

In these communities, the Stable Diffusion~\cite{rombach2021highresolution} community is currently one of the most popular and thriving ones, with a large number of open-source tools available for free, including model checkpoints finetuned on various specialized datasets. Additionally, various LoRAs~\cite{hu2022lora} and ControlNets~\cite{zhang2023adding} are available in these communities for efficient control over the results generated by Stable Diffusion. LoRAs achieve control by efficiently modifying some network parameters in a low-rank way, while ControlNets introduce an additional network to modify the intermediate features of Stable Diffusion for control.

\section{Methods}

To address the issue of poor image reconstruction performance before and after the DDIM inversion process in diffusion models, this paper employs the Enhance-Flow model as a supplement to the coverless steganography process. In Fig.\ref{fig:frame}, leveraging the reversible nature and efficient distribution transformation capability of flow models, it constructs a mapping from the original input image, Secret-image $\mathbf{x}_{s-GT}$, to the intermediate noise in the diffusion model, Container-Noise $\mathbf{y}_{n-Diff}$. The diffusion model generates the Phase1-Container $\mathbf{y}_{c-Diff}$ from Container-Noise $\mathbf{y}_{n-Diff}$. The role of the Embed-Flow model is to hide multiple images, text data, and other metadata into the Container-Image $\mathbf{y}_{c-Diff}$, producing a second-order container image, Phase2-Container $\mathbf{y}_{c-Add}$. The Phase2-Container $\mathbf{y}_{c-Add}$ maintains robustness under various noise interferences, allowing the embedded images and text data to be decoded in reverse. Under higher levels of degradation, the Phase1-Container $\mathbf{y}_{c-Diff}$ still preserves the original information, which, through the diffusion model, generates the initial restored image $\mathbf{x}_{s-Diff}$, and through the reverse process of the flow model, produces a high-fidelity restored image ${\mathbf{x^{\prime}}_{s-GT}}$.

\vspace{2pt}
\noindent \textbf{Tier-1 Importance:} For images of primary importance, you use the diffusion model to generate container images for concealment. This method is robust against a wide range of complex degradations, including transformations by deep neural networks. It ensures the security and integrity of the most critical images under various challenging conditions.

\vspace{2pt}
\noindent \textbf{Tier-2 Importance:} For images of secondary importance, you employ a second-level Embed-Flow for robust, high-capacity image steganography. This level is resistant to common degradations like noise and compression. However, it might be less effective against severe and unknown types of degradation.

\subsection{Diffusion-based Image Inversion}

Our basic framework HIS is based on a conditional diffusion model, whose noise estimator is represented by $\boldsymbol{\epsilon}_{\boldsymbol{\theta}}$, and two different conditions that serve as inputs to the diffusion model. In our work, these two conditions can serve as the private key $\mathbf{k}_{pri}$ and public key (denoted as $\mathbf{k}_{pub}$, as shown in Fig.\ref{fig:ddim-inversion}, with detailed workflow described in Algo.\ref{alg:hide workflow} and Algo.~\ref{alg:decode workflow}.
We will introduce the entire Dif framework in two parts: the hide process and the reveal process.

We first implement the Secret-Image to Container-Image transformation through the Diffusion models, which is realized by DDIM Inversion. Secret-Image is guided by Prompt1 to produce Container-Noise, and Container-Noise is guided by Prompt2 to produce Container-Noise, which is guided by Prompt2 to produce Container-Noise. Container-Image is generated under the guidance of Prompt2.

\vspace{1pt}
\textbf{The Process of Hiding Stage. } In the hiding stage, we attempt to perform translation between the secret image $\mathbf{x}_{s-Tier1}$ and the container image $\mathbf{x}_{c-Diff}$ using the forward and backward processes of deterministic DDIM.
In order to make the images before and after the translation different, we use the pre-trained conditional diffusion model with different conditions in the forward and backward processes respectively. 
These two different conditions also serve as private and public keys in the HIS framework.
Specifically, the private key $\mathbf{k}_{pri}$ is used for the forward process, while the public key $\mathbf{k}_{pub}$ is used for the backward process. After getting the container image $\mathbf{x}_{c-Diff}$, it will be transmitted over the Internet and publicly accessible to all potential receivers. 

\vspace{1pt}
\textbf{The Roles of the \textit{Private} and \textit{Public} Keys} In the proposed HIS, we found that these given conditions can act as keys in practical use. The private key is used to describe the content in the secret image, while the public key is used to control the content in the container image. For the public key, it is associated with the content in the container image, so even if it is not manually transmitted over the network, the receiver can guess it based on the received container image (described in Scenario\#2 of Fig.~\ref{fig:multi-distort}). For the private key, it determines whether the receiver can successfully reveal the original image, so it cannot be transmitted over public channels.

\begin{algorithm}[t]
   \caption{The Hide Process of Diffusion-based Inversion.}
   \label{alg:hide workflow}
\begin{algorithmic}
   \STATE {\bfseries Input:} The secret image $\mathbf{x}_{s-Tier1}$ which will be hidden, a pre-trained conditional diffusion model with noise estimator $\boldsymbol{\epsilon}_{\boldsymbol{\theta}}$, the number $T$ of time steps for sampling and two different conditions $\mathbf{k}_{pri}$ and $\mathbf{k}_{pub}$ which serve as the private and public keys.
   \STATE {\bfseries Output: } The container image $\mathbf{x}_{c-Diff}$ used to hide the secret image $\mathbf{x}_{s-Tier1}$.

   \STATE $\mathbf{y}_{n-Diff} = \mathrm{ODESolve}(\mathbf{x}_{s-Tier1}; \boldsymbol{\epsilon}_{\boldsymbol{\theta}}, \mathbf{k}_{pri}, 0, T)$  %\quad // convert image domain data to noise domain
   \STATE $\mathbf{y}_{c-Diff} = \mathrm{ODESolve}(\mathbf{y}_{n-Diff}; \boldsymbol{\epsilon}_{\boldsymbol{\theta}}, \mathbf{k}_{pub}, T, 0)$  %\quad // convert noise domain data to other image domain

   \STATE {\bfseries return}  $\mathbf{x}_{c-Diff}$

\end{algorithmic}
\end{algorithm}

\begin{algorithm}[t]
   \caption{The Reveal Process of Diffusion-based Inversion.}
   \label{alg:decode workflow}
\begin{algorithmic}
    \STATE {\bfseries Input:} The container image ${\mathbf{y}^{\prime}_{c-Diff}}$ that has been transmitted over the Internet (may be degraded from $\mathbf{y}_{c-Diff}$), the pre-trained conditional diffusion model with noise estimator $\boldsymbol{\epsilon}_{\boldsymbol{\theta}}$,  the number $T$ of time steps for sampling, the private key $\mathbf{k}_{pri}$ and public key $\mathbf{k}_{pub}$.
   \STATE {\bfseries Output: } The revealed image $\mathbf{x}_{s-Diff}$.
   
   \STATE ${\mathbf{y^{\prime}}_{n-Diff}} = \mathrm{ODESolve}({\mathbf{y^{\prime}}_{c-Diff}}; \boldsymbol{\epsilon}_{\boldsymbol{\theta}}, \mathbf{k}_{pub}, 0, T)$  %\quad // convert image domain data to noise domain
   \STATE $\mathbf{x}_{s-Diff} = \mathrm{ODESolve}({\mathbf{y^{\prime}}_{n-Diff}}; \boldsymbol{\epsilon}_{\boldsymbol{\theta}}, \mathbf{k}_{pri}, T, 0)$  %\quad // convert noise domain data to other image domain

   \STATE {\bfseries return}  $\mathbf{x}_{s-Diff}$
\end{algorithmic}
\end{algorithm}

\vspace{1pt}
\textbf{The Process of Revealing Stage. } In the reveal stage, assuming that the container image has been transmitted over the Internet and may have been damaged as ${\mathbf{x^{\prime}}_{c-Diff}}$, the receiver needs to reveal it back to the secret image through the same forward and backward process using the same conditional diffusion model with corresponding keys. Throughout the entire coverless image steganography process, we do not train or fine-tune the diffusion models specifically for image steganography tasks but rely on the inherent invertible image translation guaranteed by the DDIM Inversion.

\subsection{Enhance-Flow}

\subsubsection{\textbf{Framework}}

\begin{figure}[t]
    \centering
\includegraphics[width=0.98\linewidth]{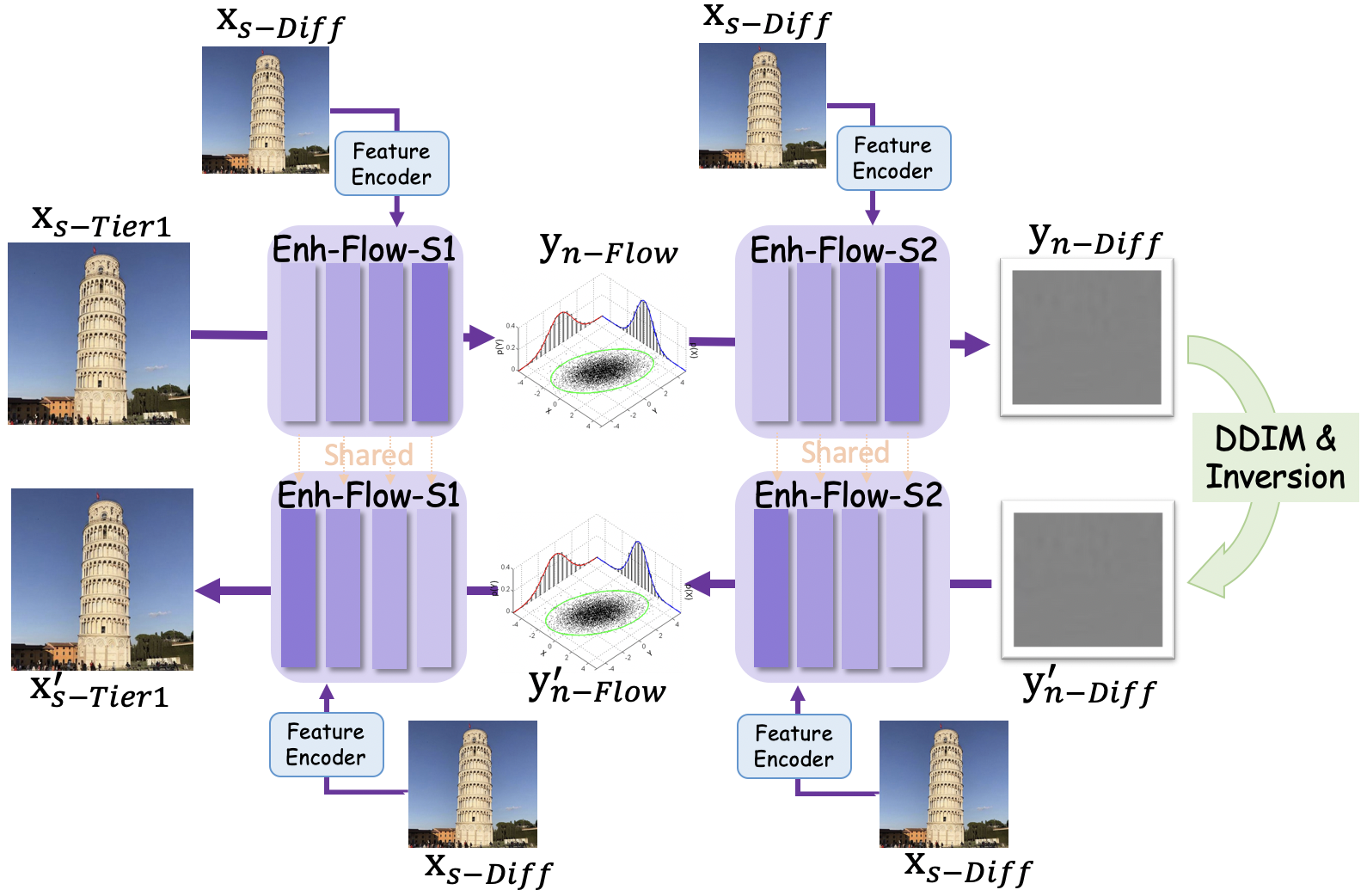}
    \caption{Framework of Enhance-flow: $\mathbf{x}_{s-Tier1}$ is firstly transfered to Gaussian-like $\mathbf{y}_{n-Flow}$ via Enh-Flow-S1 with the feature of $\mathbf{x}_{s-Diff}$, followed by the conversion to $\mathbf{y}_{n-Diff}$ via Enh-Flow-S2. The reversion from $\mathbf{y}^{\prime}_{n-Diff}$ to $\mathbf{x}^{\prime}_{s-Tier1}$ is symmetrical to the forward process.
    }
    \vspace{-0.4cm}
    \label{fig:enhflow}
\end{figure}

The diffusion model exhibits excellent generative capabilities, producing diverse $\mathbf{y}_{c-Diff}$ results different from $\mathbf{x}_{s-GT}$ under text guidance, and demonstrates strong robustness. However, its restoration performance is limited, only able to generate a roughly similar $\mathbf{x}_{s-Diff}$ from $\mathbf{y}_{c-Diff}$. To address this, the paper introduces an Enhance-Flow, aiming to generate higher quality restorations guided by the low-quality $\mathbf{x}_{s-Diff}$ results. In Fig.\ref{fig:enhflow}, Enhance-Flow consists of two flow model stages. During the Hiding phase, Stage-1 flow model uses $\mathbf{x}_{s-GT}$ as a guiding condition to transform the input image into a Gaussian distribution $\mathbf{y}_{n-flow}$. Similarly, in Stage-2, guided by $\mathbf{x}_{s-GT}$, the noise of approximate Gaussian distribution is transformed into $\mathbf{y}_{n-Diff}$ for subsequent diffusion model input.

In the Revealing phase, the diffusion model first restores ${\mathbf{y^{\prime}}_{c-Diff}}$ into ${\mathbf{y^{\prime}}_{n-Diff}}$ under public key guidance, and then reconstructs it into $\mathbf{x}_{s-Diff}$ under private key guidance, providing a lower-quality result sufficient for the flow model's auxiliary condition. Stage-2 flow model takes ${\mathbf{y^{\prime}}_{n-Diff}}$ as input and $\mathbf{x}_{s-Diff}$ as an auxiliary guiding condition, first reconstructing $\mathbf{y^{\prime}}_{n-Flow}$, and then Stage-1 flow model uses $\mathbf{y^{\prime}}_{n-Flow}$ under $\mathbf{x}_{s-Diff}$ guidance to produce the restored result $\mathbf{x^{\prime}}_{s-GT}$.

\subsubsection{\textbf{Training Protocol}}

\paragraph{\textbf{Enhanceflow-S1 }} Enhanceflow-S1 is solely trained in the first phase, with the goal of high-quality reconstruction of the original input secret image's details from a Gaussian distribution. Supposing $\mathbf{x}_{s-GT}$ serves as the model input under the condition of $\mathbf{x}_{s-Diff}$, it is transformed to an approximate Gaussian distribution $\mathbf{x}_{n-Flow}$ via EnhanceFlow-S1. The loss function for this training process is the Cross-Entropy distance between $\mathbf{y}_{n-Flow}$ and the Gaussian distribution.

\begin{align}
& \mathcal{L}_{distr-S1} = \ell_{\mathcal{CE}}(p(\mathbf{x}_{n-Flow}),\mathcal{N}(\mathbf{0},\mathbf{I})).
\label{eq.distr-s1}
\\
& \mathcal{L}_{norm-S1} = ||{\mathbf{x^{\prime}}_{s-GT}} -\mathbf{x}_{s-GT}||_2.
\label{eq.norm-s1}
\end{align}

In the reverse process, we directly sample from the Gaussian distribution to obtain $\mathbf{x^{\prime}}_{n-Flow}$. Under the supervision of $\mathbf{x}_{s-Diff}$, this is then transformed back to $\mathbf{x^{\prime}}_{s-GT}$ through the reverse process of the shared-weight EnhanceFlow-S1. This process is supervised using an L2 Loss.

\paragraph{\textbf{EnhanceFlow-S2}} We independently train the second stage of Enhanceflow-S2, aiming to compress redundant information from the diffusion model's intermediate noise $\mathbf{x^{\prime}}_{n-Diff}$ under the condition of $\mathbf{x}_{s-Diff}$, to obtain an intermediate noise $\mathbf{x^{\prime}}_{n-Flow}$ that approximates a Gaussian distribution. This noise retains key information beneficial for image reconstruction. This process is constrained using the Cross-Entropy distance between the intermediate noise $\mathbf{x^{\prime}}_{n-Flow}$ and the Gaussian distribution.

\begin{figure}[t]
 \vspace{-5pt}
  \centering
 
 % \fbox{\rule{0pt}{0.5in} \rule{0.9\linewidth}{0pt}}
  \includegraphics[width=0.97\linewidth]{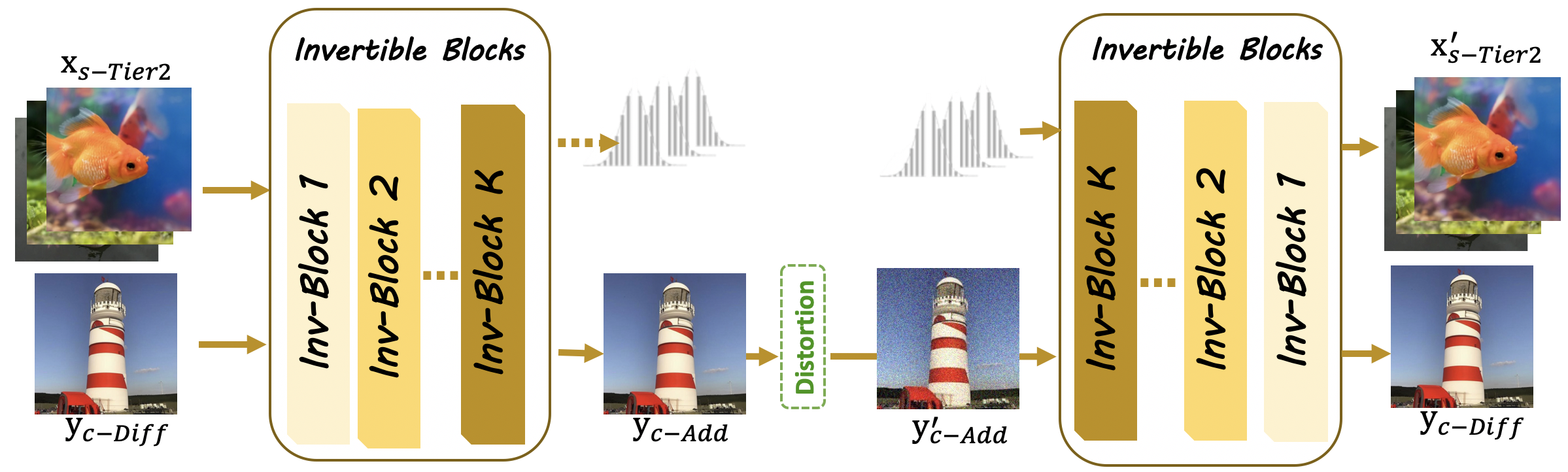}
  \vspace{-7pt}
   \caption{Framework of Embed-Flow: the Invertible Blocks take the $\mathbf{y}_{c-Diff}$ and $\mathbf{x}_{s-Tier2}$ as input to produce container $\mathbf{y}_{c-Add}$ and redundant Gaussian Noise. In the backward pass, Gaussian sampling datapoint and distorted $\mathbf{y}^{\prime}_{c-Add}$ will produce recovered secret $\mathbf{x}^{\prime}_{s-Tier2}$.}
   \label{fig:embedflow}
\end{figure}

 % \begin{wraptable}{r}{7cm}
\begin{table}[t]
% \vspace{-0.4cm}
\footnotesize
  \newcommand{\tabincell}[2]{\begin{tabular}{@{}#1@{}}#2\end{tabular}}
  \centering
  \resizebox{1.\linewidth}{!}
  {
  \begin{tabular}{l|c|c|c|c}
  \toprule[1pt]
  Methods
  % & \multirow{2}{*}[-2pt]{\tabincell{c}{Pre-defined Cover $\uparrow$\\}}
  & clean
  % & \multicolumn{3}{c}{$\lvert$Detection Accuracy - 50$\rvert$ $\downarrow$}
  & JPEG Q=80
& JPEG Enhancer
& Guassian $\sigma$=10
\\
  % \midrule
  % \toprule[1pt]
  \hline
  \hline
  % \multirow{2}{*}{Baluja}
  % & 1.00 &  98.21 & 69.23 & 100.00 & 100.00 & 100.00 \\
  % & 0.25 & 100.00 & 68.26 &  97.37 &  99.56 & 100.00 \\
  % \midrule
  % \multirow{2}{*}{ICMR}
  % & 1.00 &  89.95 & 75.96 & 98.31 & 97.62 & 99.04 \\
  % & 0.25 & 100.00 & 67.30 & 97.56 & 96.37 & 98.80 \\

\rowcolor{gray!20}
1-Tier2  & 28.39 & 23.49 & 22.05 & 24.23  \\
  % \midrule
2-Tier2  & 23.04 & 22.05 & 21.11 & 22.34  \\
  % \midrule
3-Tier2  & 24.09 & 22.78 & 20.87 & 21.84 \\
  % \midrule
4-Tier2  & 22.56 & 19.44 & 18.56 & 20.31  \\
  % \midrule
  \bottomrule[1pt]
  \end{tabular}
  }
  \caption{Average revealed PSNR with different number of embedded images $\mathbf{x}_{s-Tier2}$ via Embed-Flow under distortions. The recovery quality degrades when embedding multiple $\mathbf{x}_{s-Tier2}$ images, but can still resist various distortion.}
  \vspace{-15pt}
  \label{tab:embed-ab}
  \end{table}

\begin{align}
& \mathcal{L}_{distr-S2} = \ell_{\mathcal{CE}}(p({\mathbf{y^{\prime}}_{n-Flow}},\mathcal{N}(\mathbf{0},\mathbf{I})).
\label{eq.distr-s2}
\\
& \mathcal{L}_{norm-S2} = ||{\mathbf{y^{\prime}}_{n-Diff}} -\mathbf{y}_{n-Diff}||_2.
\label{eq.norm-s2}
\end{align}

In the reverse process of Enhanceflow-S2, we directly sample from the Gaussian distribution to get $\mathbf{y}_{n-Flow}$. Similarly, under the supervision of $\mathbf{x}_{s-Diff}$, it is transformed into $\mathbf{x^{\prime}}_{s-GT}$ through the reverse process of EnhanceFlow-S2 with shared weights. We supervise this process with an L2 Loss.

\paragraph{\textbf{Jointly Training}}
Then, we train a model holistically, where the input is ${\mathbf{y^{\prime}}_{n-Diff}}$. Under the condition of ${\mathbf{x}_{s-Diff}}$, it transforms into an approximate Gaussian $\mathbf{y}_{n-Flow}$. Subsequently, this intermediate approximate Gaussian noise is restored under the supervision of $\mathbf{x}_{s-Diff}$ into a higher-quality ${\mathbf{x^{\prime}}_{s-GT}}$.

In this training process, the primary constraint is the L2 distance between $\mathbf{x}_{s-GT}$ and the original image ${\mathbf{x^{\prime}}_{s-GT}}$. On the other side, it's to ensure that the intermediate noise $\mathbf{y_{n-Flow}}$ closely approximates a Gaussian distribution.

\begin{align}
& \mathcal{L}_{distr-joint} = \ell_{\mathcal{CE}}(p(\mathbf{y}_{n-Flow}),\mathcal{N}(\mathbf{0},\mathbf{I})).
\label{eq.distr-joint}
\\
& \mathcal{L}_{norm-joint} = ||{\mathbf{x^{\prime}}_{s-GT}} -\mathbf{x}_{s-GT}||_2.
\label{eq.norm-joint}
\end{align}

\paragraph{\textbf{Inference Process}}

During the inference testing period, in the forward process, the input image is processed through the diffusion model under the control of the private key to obtain $\mathbf{y}_{n-Diff}$. It is then transformed into the publicly disseminated container image $\mathbf{y}_{c-Diff}$ under the control of the public key through DDIM inversion. During its public network transmission, it may undergo various degradations and interferences, or be embedded with a second-level secret information $\mathbf{x}_{Add}$.

At the decoding end, the degraded container image $\mathbf{y}^{\prime}_{c-Diff}$ is received, and the intermediate noise $\mathbf{y}_{n-Diff}$ is generated under the guidance of the public key through the diffusion model. Further, under the condition of obtaining the private key, the hidden image $\mathbf{x}_{s-Diff}$ is restored in reverse through the diffusion model.
On this basis, to obtain higher quality hidden image recovery results, EnhanceFlow-S2 is first called with $\mathbf{y}^{\prime}_{n-Diff}$ as input, extracting information from the rough recovery image of $\mathbf{x}_{s-Diff}$, transforming it to the approximate Gaussian $\mathbf{y}_{n-Flow}$, and then further restored to $\mathbf{x}^{\prime}_{s-GT}$ through EnhanceFlow-S1, thereby achieving a high-quality restored hidden image.

\begin{figure*}
\includegraphics[width=1.\linewidth]{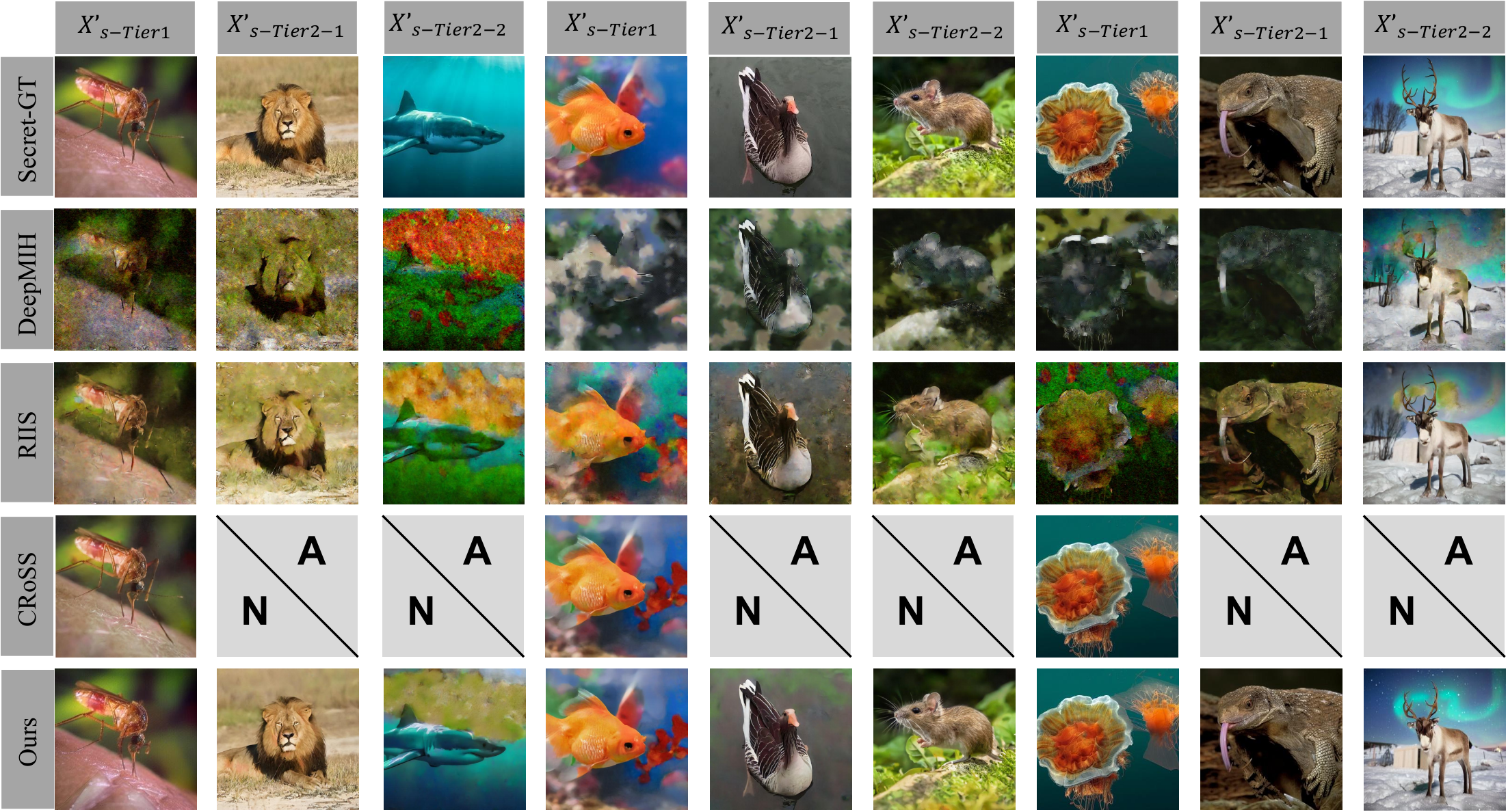}
\vspace{-20pt}
\caption{Visual results of HIS and other methods~\cite{xu2022robust, guan2022deepmih} under the same degradations of JPEG QF-80. Our method can vividly reconstruct the secret images, while other methods exhibit significant color distortion or have completely failed.}
% \Description{figure description}
\label{fig:robust-show} 

\end{figure*}

\begin{table*}[t]
\footnotesize
  \newcommand{\tabincell}[2]{\begin{tabular}{@{}#1@{}}#2\end{tabular}}
  \centering
  \resizebox{1.\linewidth}{!}
  {
  \begin{tabular}{l |c|ccc|ccc|ccc|ccc}
  \toprule[1pt]
   \multirow{2}{*}{\makecell{Methods}}
  & \multirow{2}{*}{\makecell{clean}}
  & \multicolumn{3}{c}{Gaussian noise}
  & \multicolumn{3}{c}{Gaussian denoiser~\cite{wang2021real}}
  & \multicolumn{3}{c}{JPEG compression} 
  & \multicolumn{3}{c}{JPEG enhancer~\cite{wang2021real}} \\
  % \cmidrule(l){3-5} \cmidrule(l){6-8} \cmidrule(l){9-11} \cmidrule(l){12-14}
  & & $\sigma$ = 10 & $\sigma$ = 20 & $\sigma$ = 30 & $\sigma$ = 10 & $\sigma$ = 20 & $\sigma$ = 30 & Q = 20 & Q = 40 & Q = 80 & Q = 20 & Q = 40 & Q = 80 \\
  % \midrule
  % \toprule[1pt]
  \hline
  \hline
  Baluja \cite{baluja2019hiding} & 34.24 & 10.30 & 7.54 & 6.92 & 7.97 & 6.10 & 5.49 & 6.59 & 8.33 & 11.92 & 5.21 & 6.98 & 9.88 \\
  % \midrule
  ISN \cite{lu2021large} & 41.83 & 12.75 & 10.98 & 9.93 & 11.94 & 9.44 & 6.65 & 7.15 & 9.69 & 13.44 & 5.88 & 8.08 & 11.63 \\
  % \midrule
  DeepMIH \cite{guan2022deepmih} & \textcolor{blue}{42.98} & 12.91 & 11.54 & 10.23 & 11.87 & 9.32 & 6.87 & 7.03 & 9.78 & 13.23 & 5.59 & 8.21 & 11.88 \\
  % \midrule
  RIIS \cite{xu2022robust}& \textcolor{red}{43.78} & \textcolor{red}{26.03} & \textcolor{blue}{18.89} & 15.85 & {20.89} & {15.97} & {13.92} & \textcolor{blue}{22.03} & \textcolor{blue}{25.41} & \textcolor{blue}{26.02} & {13.88} & {16.74} & {20.13} \\
  % \midrule

  CRoSS~\cite{yu2024cross} & 23.79 & {21.89} & {20.19} & \textcolor{blue}{18.77} & \textcolor{blue}{21.39} & \textcolor{blue}{21.24} & \textcolor{blue}{21.02} & {21.74} & {22.74} & {23.51} & \textcolor{blue}{20.60} & \textcolor{blue}{21.22} & \textcolor{blue}{21.19} \\
    % \midrule
\rowcolor{gray!20}
  HIS & 28.39 & \textcolor{blue}{23.49} & \textcolor{red}{21.88} & \textcolor{red}{20.02} & \textcolor{red}{21.73} & \textcolor{red}{22.41} & \textcolor{red}{21.98} & \textcolor{red}{23.21} & \textcolor{red}{26.11} & \textcolor{red}{26.23} & \textcolor{red}{22.42} & \textcolor{red}{23.23} & \textcolor{red}{23.15} \\
  \bottomrule[1pt]
  \end{tabular}
}
  \caption{PSNR(dB) results of the proposed HIS and other methods under different levels of degradations. The proposed HIS can achieve superior data fidelity in most settings. The best results are \textcolor{red}{red} and the second-best results are \textcolor{blue}{blue}.}
  \label{tab:robust}
  \vspace{-12pt}
\end{table*}

\begin{figure}[t!]
    \centering
\includegraphics[width=1.\linewidth]{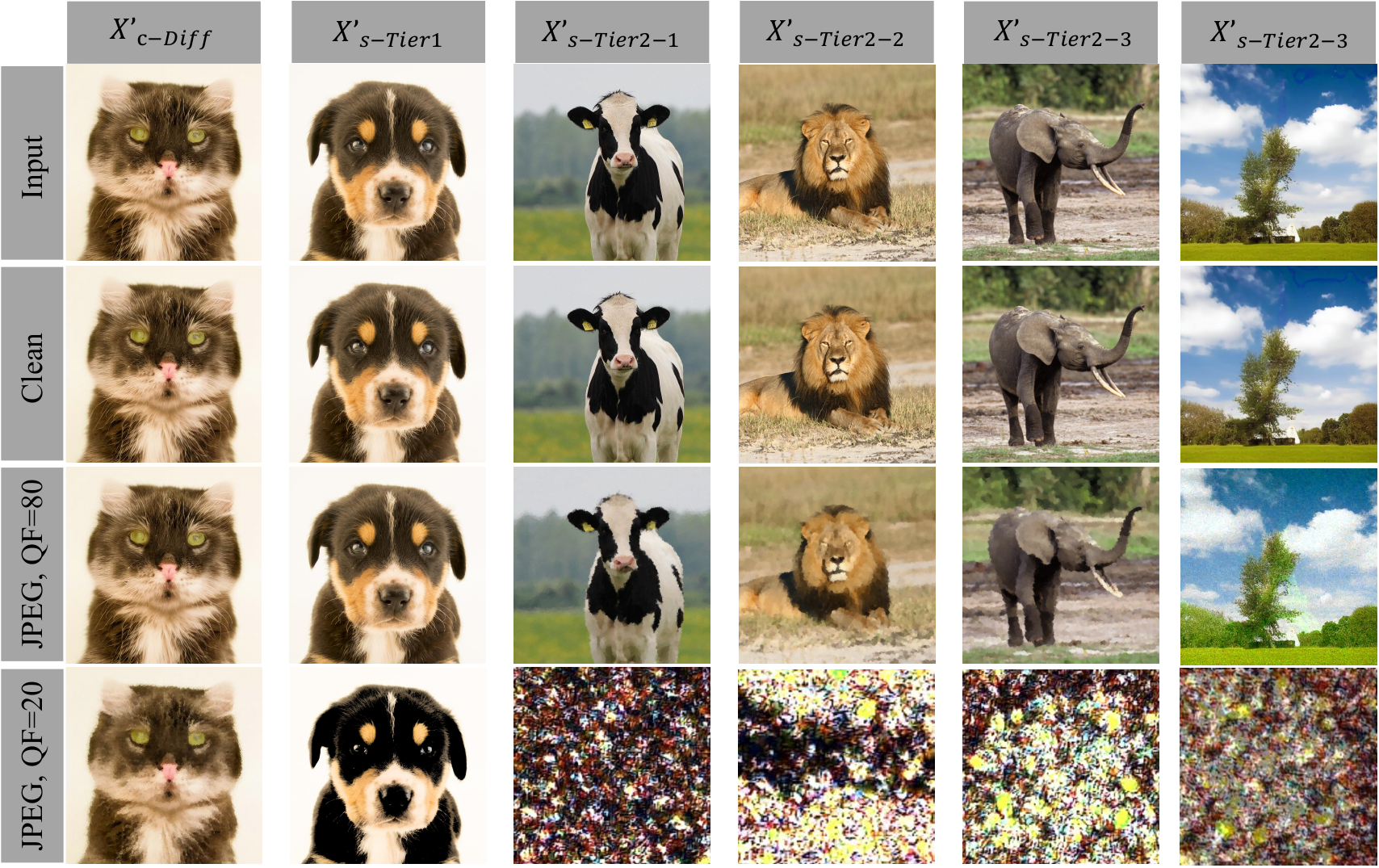}
    \vspace{-20pt}
    \caption{Revealed Results of multiple secret images (a single $\mathbf{x}^{\prime}_{s-Tier1}$ and 4 $\mathbf{x}^{\prime}_{s-Tier2}$), $\mathbf{x}^{\prime}_{s-Tier1}$ can be robustly revealed under severe distortion.
    }
    \vspace{-1pt}
    \label{fig:multi-distort}
\end{figure}

\subsection{Embed-Flow}

The container images produced by the diffusion model exhibit strong robustness, capable of withstanding varied and complex degradation conditions, even after undergoing complex transformations such as through neural networks. In the current network environment, there's a pressing need for image steganography algorithms with higher capacity and robustness. Hence, we utilize $\mathbf{y}_{c-Diff}$ from the previous phase as the carrier, designing the Embed-Flow model to conceal multiple images or large volumes of text information within it.

The specific flow model conditions of Embed-Flow are decided by the user's needs. In Fig.~\ref{fig:embedflow}, when hiding multiple frames, Embed-Flow first applies a Haar wavelet transform to the inputs ${\mathbf{y^{\prime}}_{c-Diff}}$ and $\mathbf{x}_{s-Add}$, and then uses flow model transformations with ${\mathbf{y^{\prime}}_{c-Diff}}$ as an auxiliary condition. The aim is to transform the channels corresponding to $\mathbf{x}_{s-Add}$ into an approximate Gaussian distribution, thereby hiding the information in $\mathbf{x}_{c-Add}$. The obtained $\mathbf{x}_{c-Add}$ can resist common benign degradations and exhibit broad robustness. At the receiving end, the same Embed-Flow can be used for inverse operations to decrypt and restore $\mathbf{y^{\prime}}_{c-Diff}$ and $\mathbf{x^{\prime}}_{s-Add}$. Here, $\mathbf{y^{\prime}}_{c-Diff}$ will continue as the input for the directional decoding of DDIM or Enhance-Flow in the first stage.

\paragraph{\textbf{Loss Functions}}
Concretely, since the distribution can be tractably depicted in flow-based models, CANP encourages the $p(\mathbf{z})$ to be independent from $p(\mathbf{h_f})$ and $p(\mathbf{y})$ and approximate to Gaussian distribution by distribution loss $\mathcal{L}_{distr}$ in Eq.~(\ref{eq.distr}). We depict the distribution distance by cross-entropy (CE) on $\mathbf{z}$. To guide the container image $\mathbf{y}$ to be approximately identical to the host image $\mathbf{x_h}$ both in spatial and frequency domain, we further apply fast fourier transform (FFT) to extract frequency component in Eq.~(\ref{eq.con}).

\begin{align}
& \mathcal{L}_{rev} = ||\mathbf{x}_{s} - \mathbf{\hat{x}_{s}} ||_2 + ||\mathbf{{x}_{h}} - \mathbf{\hat{x}_{h}} ||_2.
\label{eq.rev}
\\
& \mathcal{L}_{norm} = ||\mathbf{y} - \mathbf{\tilde{y}}||_2.
\label{eq.cem}
\\
& \mathcal{L}_{distr} = \ell_{\mathcal{CE}}(p(\mathbf{z}),\mathcal{N}(\mathbf{0},\mathbf{I})).
\label{eq.distr}
\\
& \mathcal{L}_{con} = ||\mathbf{x_{h}} - \mathbf{y}||_2 + ||\text{FFT}(\mathbf{x_{h}}) - \text{FFT}(\mathbf{y})||_2.
\label{eq.con}
\\
& \mathcal{L}_{ {total }} =\lambda_{1}  \mathcal{L}_{rev} + \lambda_{2} \mathcal{L}_{con} + \lambda_{3} \mathcal{L}_{norm} + \lambda_{4} \mathcal{L}_{distr},
\label{eq.loss}
%\\
\end{align}

%Their corresponding weight factors are $\lambda_1$, $\lambda_2$, $\lambda_3$ and $\lambda_4$
%introduce $L_2$ distance to push the restored $\mathbf{\hat{y}}$ approaching original $\mathbf{{y}}$:

In summary, The total loss function in Eq.~(\ref{eq.loss}) considers the following four components: embedded image revealing, container invisibility, distortion enhancement, and noise distribution distance:
%\vspace{6pt}
% \setlength{\abovedisplayskip}{-0.2cm}
% \begin{flushleft}
% \begin{equation}
%\mathcal{L}_{\text {total }} =\lambda_{1}  \mathcal{L}_{Rev} + \lambda_{2} \mathcal{L}_{Con} + \lambda_{3} \mathcal{L}_{norm} + \lambda_{4} \mathcal{L}_{\mathrm{Distr}}
% \mathcal{L}_{ {total }} =\lambda_{1}  \mathcal{L}_{Rev} + \lambda_{2} \mathcal{L}_{Con} + \lambda_{3} \mathcal{L}_{norm} + \lambda_{4} \mathcal{L}_{Distr}.
% \label{eq.loss}
% \end{equation}
% \begin{equation}
%     \mathcal{L}_{Rev} = ||\mathbf{x}_{s} - \mathbf{\hat{x}}_{s} ||_2 + ||\mathbf{{x}}_{h} - \mathbf{\hat{x}}_{h} ||_2.
% \label{eq.rev}
% \end{equation}
% \begin{equation}
%     \mathcal{L}_{norm} = ||\mathbf{y} - \mathbf{\tilde{y}}||_2.
% \label{eq.cem}
% \end{equation}
% \begin{equation}
% \mathcal{L}_{Dist} = \ell_{\mathcal{CE}}(p(\mathbf{z}),\mathcal{N}(\mathbf{0},\mathbf{I})).
% \label{eq.distr}
% \end{equation}
% \begin{equation}
% \mathcal{L}_{Con} = ||\mathbf{x}_{h} - \mathbf{y}||_2 + ||FFT(\mathbf{x}_{h}) - FFT(\mathbf{y})||_2.
% \label{eq.con}
% \end{equation}
% \end{flushleft}

% \begin{wraptable}{r}{7cm}

\section{Experimental Results}
\subsection{Implementation and Setup Details}
Our proposed framework successfully maintains the payload capacity by hiding multiple images in one container image. HIS is implemented with the NVIDIA 3090 GPU for acceleration. We implement the Adam optimizer with $\beta_{1} = 0.9$ and $\beta_{2} = 0.99$. The learning rate is set to be 0.0001, and the batch size is set to be 16 for training. The dataset for training and testing is DIV2K \cite{DIV2K}, if not specified. For the loss, the corresponding weight factors are $\lambda_1=1$, $\lambda_2=16$, $\lambda_3=1$ and $\lambda_4=0.5$. The PSNR (Peak Signal to Noise Ratio) metric is utilized to evaluate the performance. 

\subsection{Abalation Studies}

In Fig.~\ref{fig:multi-distort}, results from HIS demonstrate its Tiered robustness, especially under severe distortions. Specifically, the method successfully recovers the highly important $\mathbf{x}^{\prime}{s-Tier1}$ image, while also managing to reveal multiple $\mathbf{x}^{\prime}{s-Tier2}$ images. This confirms that our HIS method maintains the integrity of crucial Tier-1 embedded images even in challenging conditions.

Tab.\ref{tab:embed-ab} illustrates the average revealed PSNR with varying numbers of embedded images $\mathbf{x}_{s-Tier2}$ via Embed-Flow under different distortions provides critical insights into the capabilities of our Hierarchical Image Steganography (HIS) method. It demonstrates the high capacity of HIS, affirming its ability to embed multiple images while maintaining significant robustness. Despite the expected degradation in recovery quality with an increased number of embedded images, the system effectively resists various distortions, ensuring the integrity and quality of the recovered images.

In Tab.\ref{tab:ab-flow}, to validate the effectiveness and necessity of the two-stage flow model enhancement, Enhance-Flow, proposed in this paper, we consider two simpler network architectures as alternative post-processing enhancement methods for the diffusion model's hiding-recovery process.

Case 1: Directly train the two-stage Enhance-Flow without the intermedia constraint of ${\mathbf{y}_{n-Flow}}$.

Case 2: A single-stage flow model structure, equivalent to Enhance-Flow-S2, is directly trained to map from ${\mathbf{x^{\prime}}_{s-GT}}$ to $\mathbf{y}_{n-Flow}$. T
In the forward process, the output ${\mathbf{y^{\prime}}_{n-Diff}}$ is compared with the actual noise $\mathbf{y_{n-Diff}}$ produced by the diffusion model to calculate the distance: $\mathcal{L}_{norm} = ||{\mathbf{y^{\prime}}_{n-Diff}}-\mathbf{y}_{n-Diff}||_2.$ In the reverse process, the output ${\mathbf{x^{\prime}}_{s-GT}}$ is compared with the original input image to calculate the L2 distance. $
\mathcal{L}_{norm} = ||{\mathbf{x^{\prime}}_{s-GT}} -\mathbf{x}_{s-GT}||_2.$

% \label{eq.yndiff}
% \end{equation}

% \begin{equation}
% \label{eq.x-gt}
% \end{equation}

Case 3: Train an image-to-image transformation \textbf{Unet}, inputting the noise from the diffusion model, $\mathbf{x_{s-Diff}}$, and outputting the reconstructed Secret Image ${\mathbf{x^{\prime}}_{s-GT}}$. This process is constrained using the L2 distance.

% \begin{equation}
% \mathcal{L}_{norm} = ||{\mathbf{x^{\prime}}_{s-GT}} -\mathbf{x}_{s-GT}||_2.
% \label{eq.x-gtnorm}
% \end{equation}

\begin{table}[t]
% \vspace{-0.4cm}
\footnotesize
  \newcommand{\tabincell}[2]{\begin{tabular}{@{}#1@{}}#2\end{tabular}}
  \centering
  \resizebox{1.\linewidth}{!}
  {
  \begin{tabular}{l|c|c|c|c}
  \toprule[1pt]
  Methods
  % & \multirow{2}{*}[-2pt]{\tabincell{c}{Pre-defined Cover $\uparrow$\\}}
  & clean
  % & \multicolumn{3}{c}{$\lvert$Detection Accuracy - 50$\rvert$ $\downarrow$}
  & JPEG Q=80
& JPEG Enhancer
& Guassian $\sigma$=10
\\
  % \midrule
  % \toprule[1pt]
  \hline
  \hline
  % \multirow{2}{*}{Baluja}
  % & 1.00 &  98.21 & 69.23 & 100.00 & 100.00 & 100.00 \\
  % & 0.25 & 100.00 & 68.26 &  97.37 &  99.56 & 100.00 \\
  % \midrule
  % \multirow{2}{*}{ICMR}
  % & 1.00 &  89.95 & 75.96 & 98.31 & 97.62 & 99.04 \\
  % & 0.25 & 100.00 & 67.30 & 97.56 & 96.37 & 98.80 \\

\rowcolor{gray!20}
EnhFlow  & 28.39 & 23.49 & 22.05 & 24.23  \\
  % \midrule
  % \rowcolor{gray!20}
  EnhFlow(Direct)  & 23.04 & 22.05 & 21.11 & 22.34  \\
  % \midrule
  Conditional Flow  & 24.09 & 22.78 & 20.87 & 21.84 \\
  % \midrule
  UNet  & 22.56 & 19.44 & 18.56 & 20.31  \\
  % \midrule
  \bottomrule[1pt]
  \end{tabular}
  }
  \caption{Ablation studies of different schemes of post-enhancement over DDIM inversion under distortions.}
  \label{-20pt}
  \label{tab:ab-flow}
  \vspace{-10pt}
  \end{table}
  
\begin{figure}[t]
 \vspace{-5pt}
  \centering
  \includegraphics[width=0.92\linewidth]{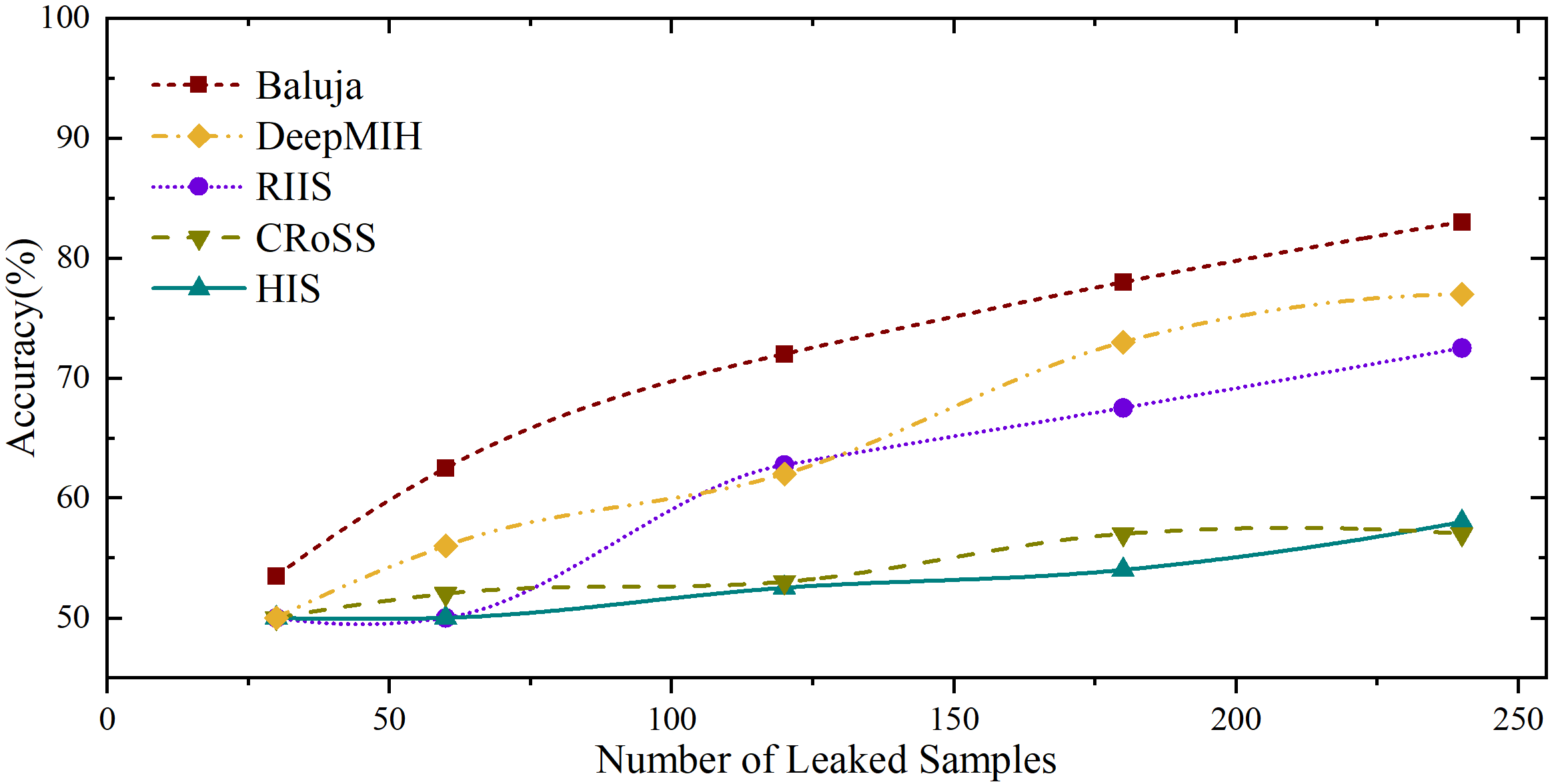}
  \vspace{-7pt}
   \caption{learning curve of steganalysis on recent steganography methods.}
   \label{fig:learn-analys}
   \vspace{-10pt}
\end{figure}

\begin{figure}[t]
 % \vspace{-5pt}
  \centering
 
 % \fbox{\rule{0pt}{0.5in} \rule{0.9\linewidth}{0pt}}
  \includegraphics[width=0.94\linewidth]{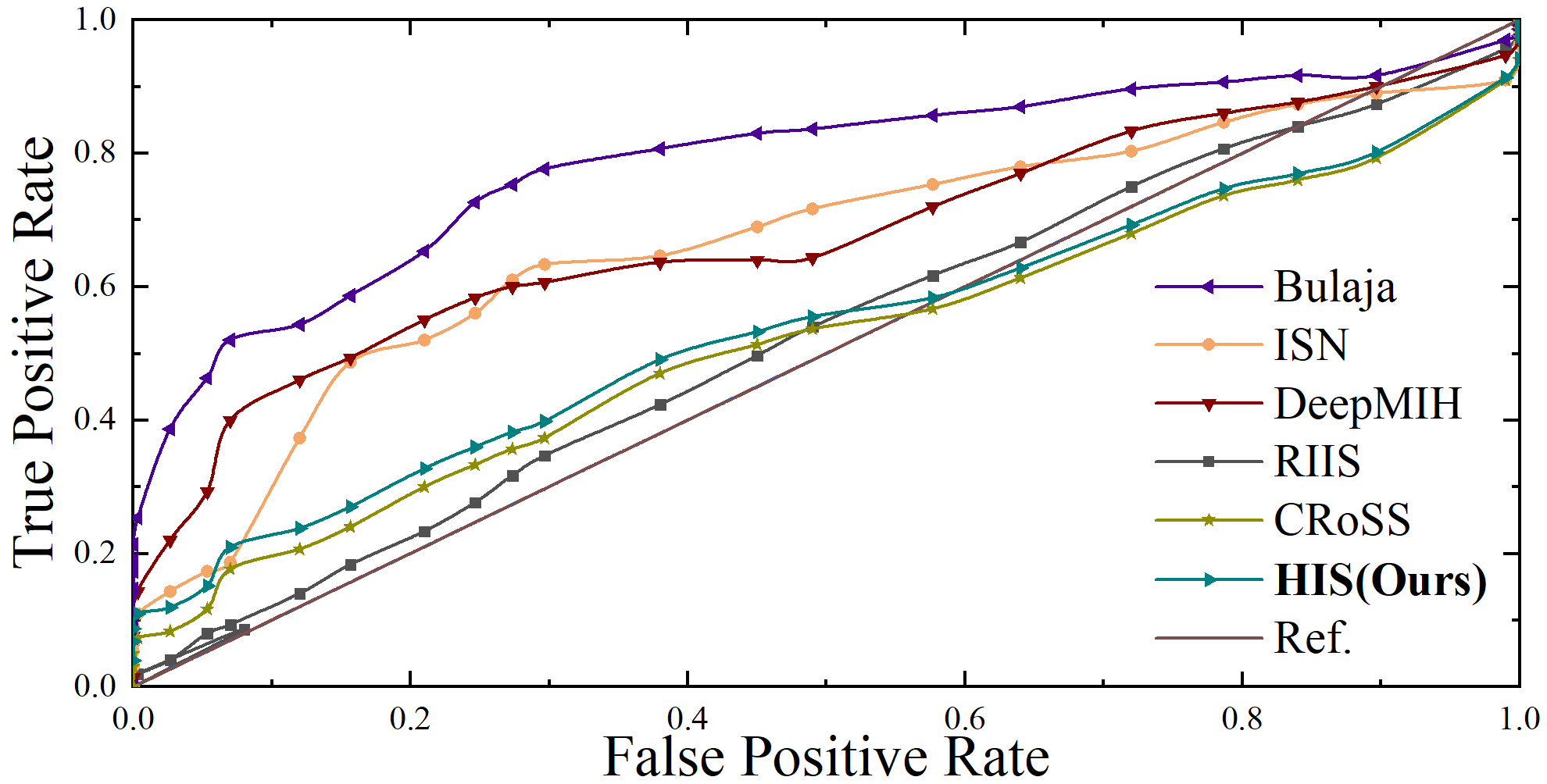}
  \vspace{-7pt}
   \caption{ROC curve of steganalysis on competitive methods. The more area under the ROC curve (auROC) approaches 0.5 (as Ref.), the less susceptible to steganalysis the method is.}
   \label{fig:roc-under}
   \vspace{-8pt}
\end{figure}
  
  \begin{figure*}[tp]  \includegraphics[width=1.\textwidth]{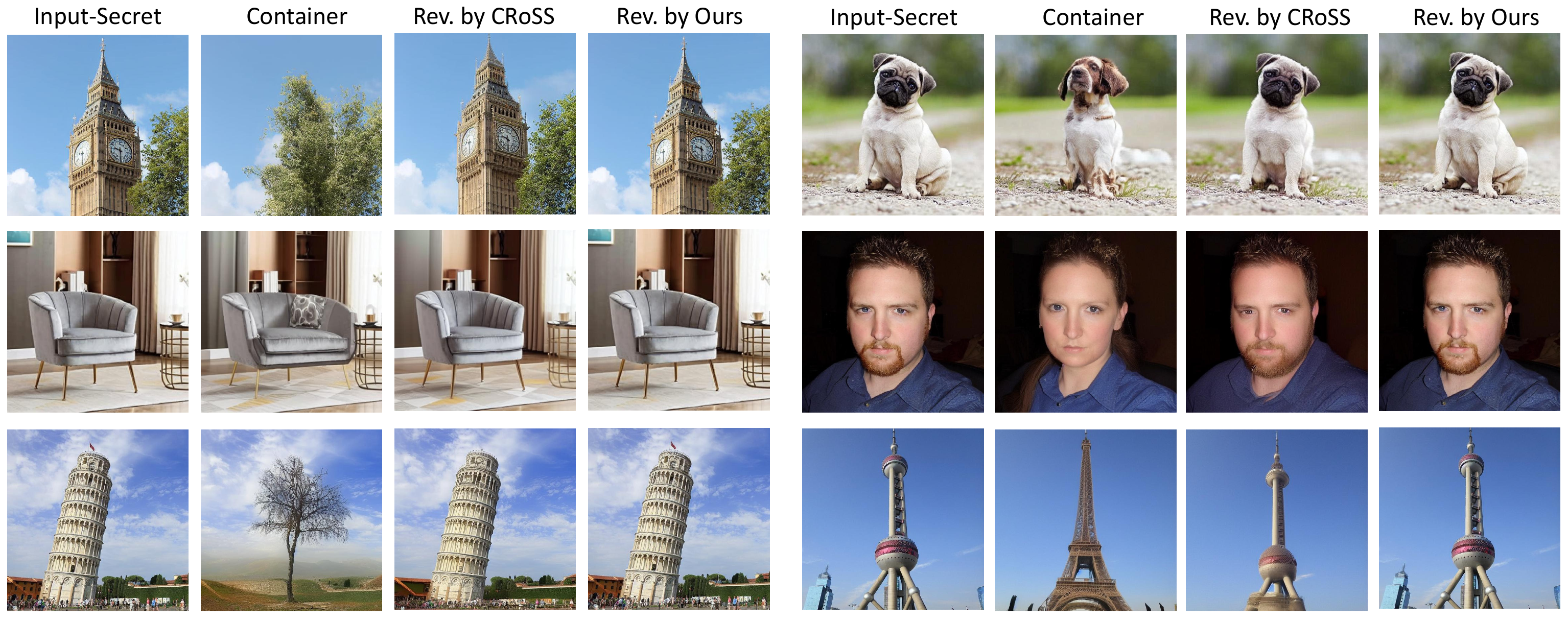}
  \vspace{-5pt}
    \captionsetup{type=figure}
    \caption{Visual results of proposed HIS: given an input $\mathbf{x}_{s-Tier1}$ \textbf{Input-Secret} to be hide, the \textbf{Container} is produced by diffusion-based image inversion, and reversely \textbf{Rev. by CRoSS} is the reconstructed secret image by the former coverless steganography method CRoSS~\cite{yu2024cross}, \textbf{Rev. by Ours} is the better-reconstructed secret by HIS.}
    \Description{figure description}
    \vspace{-0.2cm}
    \label{fig:contain-our-cross}
\end{figure*}

\subsection{Comparison with SOTA}
In the image steganography task, the most common concern is the fidelity of two pairs: revealed secret $\mathbf{\hat{x}_s}$ and origin $\mathbf{x_s}$, container $\mathbf{y}$ and host image $\mathbf{x_h}$. For the comparison with the latest method, we reproduce the State-of-the-art ISN \cite{lu2021large} and reach the performance itself claimed on the DIV2K \cite{DIV2K}. Despite the variety of colors and structures of the images, HIS can restore them with no viewable artifacts. The performance of hiding images with the container image stained by noise or JPEG compression is shown in Tab.~\ref{tab:robust}. The results reveal that our proposed method HIS successfully maintains higher reconstruction quality compared with the latest methods. To prove the payload capacity of our method, we increase the channel of HIS for hiding multiple secret images into one container. 

In Fig.~\ref{fig:robust-show}, the visual comparisons of our HIS with others under JPEG QF-80 degradation clearly demonstrate the superior recovery capabilities of HIS. While competing methods suffer from notable color distortions or failure, HIS effectively reconstructs secret images with remarkable fidelity and robustness to compression. Fig.~\ref{fig:multi-distort} shows the model performance for hiding single or multiple secret images into one container under different distortions.

\begin{table}[t]
% \vspace{-0.4cm}
\footnotesize
  \newcommand{\tabincell}[2]{\begin{tabular}{@{}#1@{}}#2\end{tabular}}
  \centering
  \resizebox{1\linewidth}{!}
  {
  \begin{tabular}{l|c|c|c|c}
  \toprule[1pt]
  \multirow{2}{*}{\makecell{Methods}}
  % & \multirow{2}{*}[-2pt]{\tabincell{c}{Pre-defined Cover $\uparrow$\\}}
  & \multirow{2}{*}{\makecell{NIQE $\downarrow$}}
  & \multicolumn{3}{c}{$\lvert$Detection Accuracy - 50$\rvert$ $\downarrow$} \\
  % \cmidrule(l){3-5}
  & & XuNet~\cite{xu2016structural} & YedroudjNet~\cite{yedroudj2018yedroudj} & KeNet~\cite{you2020siamese} \\
  % \midrule
  % \toprule[1pt]
  \hline
  \hline
  % \multirow{2}{*}{Baluja}
  % & 1.00 &  98.21 & 69.23 & 100.00 & 100.00 & 100.00 \\
  % & 0.25 & 100.00 & 68.26 &  97.37 &  99.56 & 100.00 \\
  % \midrule
  % \multirow{2}{*}{ICMR}
  % & 1.00 &  89.95 & 75.96 & 98.31 & 97.62 & 99.04 \\
  % & 0.25 & 100.00 & 67.30 & 97.56 & 96.37 & 98.80 \\
  Baluja \cite{baluja2019hiding}   & 3.43$\pm$0.08 & 45.18$\pm$1.69 & 43.12$\pm$2.18 & 46.88$\pm$2.37 \\
  % \midrule
  ISN \cite{lu2021large}  & \textcolor{red}{2.87$\pm$0.02} & 5.14$\pm$0.44 & 3.01$\pm$0.29 & 8.62$\pm$1.19 \\
  % \midrule
  DeepMIH \cite{guan2022deepmih}  & \textcolor{blue}{2.94$\pm$0.02} & 5.29$\pm$0.44 & 3.12$\pm$0.36 & 8.33$\pm$1.22 \\
  % \midrule
  RIIS \cite{xu2022robust}
   & 3.13$\pm$0.05 & \textcolor{red}{0.73$\pm$0.13} & {0.24$\pm$0.08} & {4.88$\pm$1.15}\\ 
    % \midrule
  CRoSS~\cite{yu2024cross}  & 3.04 & {1.32} & \textcolor{blue}{0.22} & \textcolor{blue}{2.11} \\
   \rowcolor{gray!20}
     % \midrule
 HIS  & 3.10 & \textcolor{blue}{1.01} & \textcolor{red}{0.18} & \textcolor{red}{1.95} \\
  \bottomrule[1pt]
  \end{tabular}
  }

  \caption{Security analysis. NIQE indicates the visual quality of container images, lower is better. The closer the detection rate of a method approximates $50\%$, the more secure the method is considered, as it suggests its output is indistinguishable from random chance. The best results are \textcolor{red}{red} and the second-best results are \textcolor{blue}{blue}.}
  \vspace{-20pt}
  \label{tab:detect-acc}
  \end{table}
% In Fig.~\ref{pic:bigvisual}, even under slight interference on container image, the secret restoration of HiNet \cite{jing2021hinet} witnesses a substantial drop in performance. It shows that the previous methods ignorant of distortion are vulnerable and fragile, limiting their application in practice. Since the performance of the original ISN model ignorant of distortion \cite{lu2021large} is too poor, we finetuned the ISN network separately for every distortion level in our experiment settings as the baseline. In contrast to the original ISN model, we name the finetuned ISN as \textbf{ISN}$^{+}$. The ISN$^+$ is also finetuned for every specific noise or compression level, but it still fails to offer satisfactory performance. 

\begin{figure}[tp]
 % \vspace{-5pt}
  \centering
 
 % \fbox{\rule{0pt}{0.5in} \rule{0.9\linewidth}{0pt}}
  \includegraphics[width=0.897\linewidth]{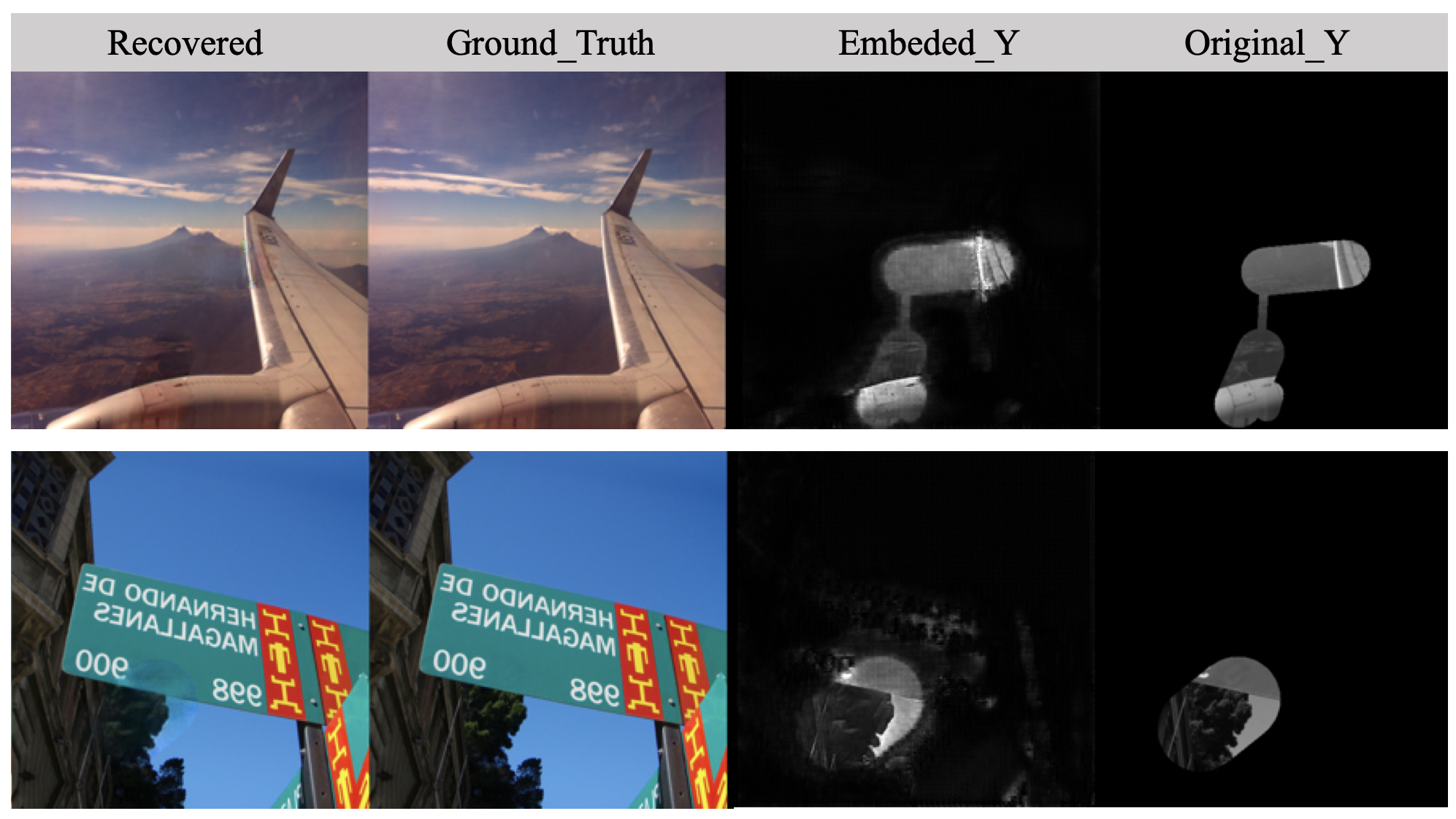}
  \vspace{-7pt}
   \caption{Examples of our applications of self-recovery and tampering localization.}
   \label{fig:immu-app}
   \vspace{-10pt}
\end{figure}

\subsection{Anti-Analysis Security}\label{exp4}

Steganalysis tools are used to identify whether an image is a container image that carries secret messages. 
Tab.~\ref{tab:detect-acc} shows the steganalysis results of different methods. Compared with existing methods, HIS successfully confused most steg-analysis tools. Our detection accuracy rates are all around 50\%, which means outstanding statistical security. Fig.~\ref{fig:roc-under} demonstrates the steganalysis performance on various methods. This implies that methods closer to this ideal auROC value are inherently more secure, as they are less likely to be identified through common steganalysis ~\cite{you2020siamese}. The auROC of HIS is most close to 0.5, suggesting a higher level of security for embedding sensitive information within images.

Deep steganalysis has been developed to investigate how many leaked training samples (\textit{i.e.}, the pair of cover and container frames) acquired by the attacker will make the Recent deep-learning methods perform better than conventional detectors. Here, we re-train the latest steg-analysis method SiaSteg \cite{you2020siamese} with leaked samples from the evaluation set. The containers are mixed with host images from DIV2K for detection of wide-range thresholds. As the Fig.~\ref{fig:learn-analys} presented, HIS reflects the lowest detection accuracy.  

HIS is designed for both the visual and statistical undetectability. Take the container images, original secret images in Fig.~\ref{fig:contain-our-cross} for example, HIS hardly arouses human-eye suspicion. 

% Additionally, we utilize a widely-used steganalysis tool [\textcolor{green}{1}] called StegExpose \cite{boehm2014stegexpose} to statistically measure the identifiability.

\subsection{Applications}

\noindent \textbf{Local Self-Recovery} Leveraging the robustness of our Hierarchical Image Steganography (HIS) method, we propose an enhanced approach for Image Self-Recovery, which is particularly effective against partial image corruption. This technology empowers digital images with the ability to auto-recover tampered sections when shared on social media, without the need for external image forgery detection or reconstruction techniques. By embedding subtle, imperceptible changes into the image, users can upload a fortified version that maintains normal usability while ensuring the integrity of the content. As Fig.\ref{fig:immu-app} shows, HIS enables the recipient to accurately localize and restore any tampered areas of the image, thereby preserving the original content with high fidelity.

\noindent \textbf{Tampering Localization} The hidden images of Tier-2 importance exhibit certain vulnerabilities, as local erasure, tampering, or replacement on the container image can lead to the loss of local information in the recovered hidden images \cite{zhang2024v2a, zhang2023editguard}. Hence, we can utilize the difference in local robustness to locate local tampering on the container image. If its local information is modified, the hidden images of Tier-2 importance will display evident defects such as dark blocks.

\noindent \textbf{Multi-Image Protection} Users on social media can securely share multiple images embedded within a single container image, utilizing HIS. Firstly, users select images and categorize them based on their sensitivity into Tier-1 (highly sensitive) and Tier-2 (less sensitive). HIS acts as an integrated tool in the social media platform embeds these images into a single normal-looking image using steganography. Tier-1 images are embedded with high robustness to protect against tampering, while Tier-2 images are embedded with less robustness. The container image is then posted publicly on the user’s profile, appearing as a typical image. 
% Only authorized users with the right tools and permissions can decode and view the sensitive content hidden within the image.
\vspace{-5pt}
\section{Conclusion}
We propose a Hierarchical Image Steganography (HIS) framework that effectively leverages the unique properties of diffusion models and flow-based models. This innovative approach achieves superior performance in terms of security, controllability, and robustness compared to existing steganography methods. By incorporating a tiered embedding strategy, HIS dynamically adjusts the robustness of embedded images based on their importance, ensuring that critical information remains intact even under significant distortions, offering a powerful tool for protecting digital content in an increasingly connected world.

% \noindent \textbf{Limitations} 

% Though our framework impresses with its robustness under complex distortion, it is still difficult to handle real-world scenarios. In addition, the conditional flow module increases the parameter and GPU memory requirement. 

% \begin{acks}
% \end{acks}

\bibliographystyle{ACM-Reference-Format}
\bibliography{main}

\end{document}